\def\year{2021}\relax
\documentclass[letterpaper]{article} 
\usepackage{aaai21}  
\usepackage{times}  
\usepackage{helvet} 
\usepackage{courier}  
\usepackage[hyphens]{url}  
\usepackage{graphicx} 
\usepackage{natbib}  
\usepackage{caption} 
\usepackage{relsize}
\usepackage{amsmath}
\usepackage[ruled]{algorithm2e}
\usepackage{tabularx,booktabs}
\usepackage{multirow}
\title{Adversarial Training with Stochastic Weight Average}
\author{
  Joong-Won Hwang \ \ \ Youngwan Lee \ \ \ Sungchan Oh  \ \ \  Yuseok Bae
}

\affiliations{
    
    \textsuperscript{\rm 1}
    Electronics and Telecommunications Research Institute \\

    218 Gajeong-ro, Yuseong-gu, Daejeon, 34129, KOREA\\
    
    \{jwhwang, ywlee, sungchan.oh, baeys\} @etri.re.kr

}
\begin{document}
\maketitle
\begin{abstract}
Adversarial training deep neural networks often experience serious overfitting problem. Recently, it is explained that the overfitting happens because the sample complexity of training data is insufficient to generalize robustness. 
In traditional machine learning, one way to relieve overfitting from the lack of data is to use ensemble methods.
However, adversarial training multiple networks is extremely expensive. Moreover, we found that there is a dilemma on choosing target model to generate adversarial examples. Optimizing attack to the members of ensemble will be suboptimal attack to the ensemble and incurs covariate shift, while attack to ensemble will weaken the members and lose the benefit from ensembling.
In this paper, we propose adversarial training with Stochastic weight average~(SWA); while performing adversarial training, we aggregate the temporal weight states in the trajectory of training. 
By adopting SWA, the benefit of ensemble can be gained without tremendous computational increment and without facing the dilemma. Moreover, we further improved SWA to be adequate to adversarial training. 
The empirical results on CIFAR-10, CIFAR-100 and SVHN show that our method can improve the robustness of models.
\end{abstract}

\section{Introduction}
\noindent Although DNN shows great performance and generalization ability, it has been found that convolutional neural networks~(CNNs) are susceptible to designed adversarial attack, even if the attack is imperceptibly small.
As the modern computer vision technology heavily relies on CNN, the vulnerability becomes a great threat. 
To mitigate such threat, many algorithms have been proposed to make network to be robust to adversarial attack~\cite{xie2018mitigating,song2018pixeldefend,papernot2016distillation,madry2017towards}.
However, it has been found that many of these methods are not robust indeed; many proposed defense algorithms rely on obfuscated gradient which gives robustness against particular  attacks only, and can be circumvented by adaptively designed attack~\cite{athalye2018obfuscated, tramer2017ensemble}. 
Among the methods compared in the paper, the only defense method that is believed to provide true robustness is training network with strong adversarial examples such as projected gradient descent (PGD)~\cite{madry2017towards}.
Hence, adversarial training is considered as the most reliable defense paradigm so far. 

However, adversarial training is prone to overfitting~\cite{schmidt2018adversarially}. 
In robust training, adversarial loss on test set increases substantially after a certain point while training error decrease continuously. 
This overfitting phenomenon brings large generalization gap on adversarial accuracy and limits robustness. 
To explain such susceptibility, Schmidt~\cite{schmidt2018adversarially} provide theoretical analysis that concludes the overfitting is due to the lack of training data. 
According to the analysis, the sample complexity required to generalize robustness is significantly lager than that for standard generalization.

 In traditional machine learning, one way to relieve overfitting is adopting ensemble methods~\cite{dietterich2000ensemble}. 
 When the hypothesis space is too large to explore for limited training data, there may be several different hypotheses giving similar accuracy on the training data, and combining the hypotheses can reduce the risk to choose wrong hypotheses.
 Therefore, to alleviate the overfitting, one can easily consider naive ensemble method which combines the results of adversarial trained model~\cite{grefenstette2018strength}.
\begin{figure*}[h]
\centering
\scalebox{1.0}{
  \includegraphics[width=0.95\textwidth]{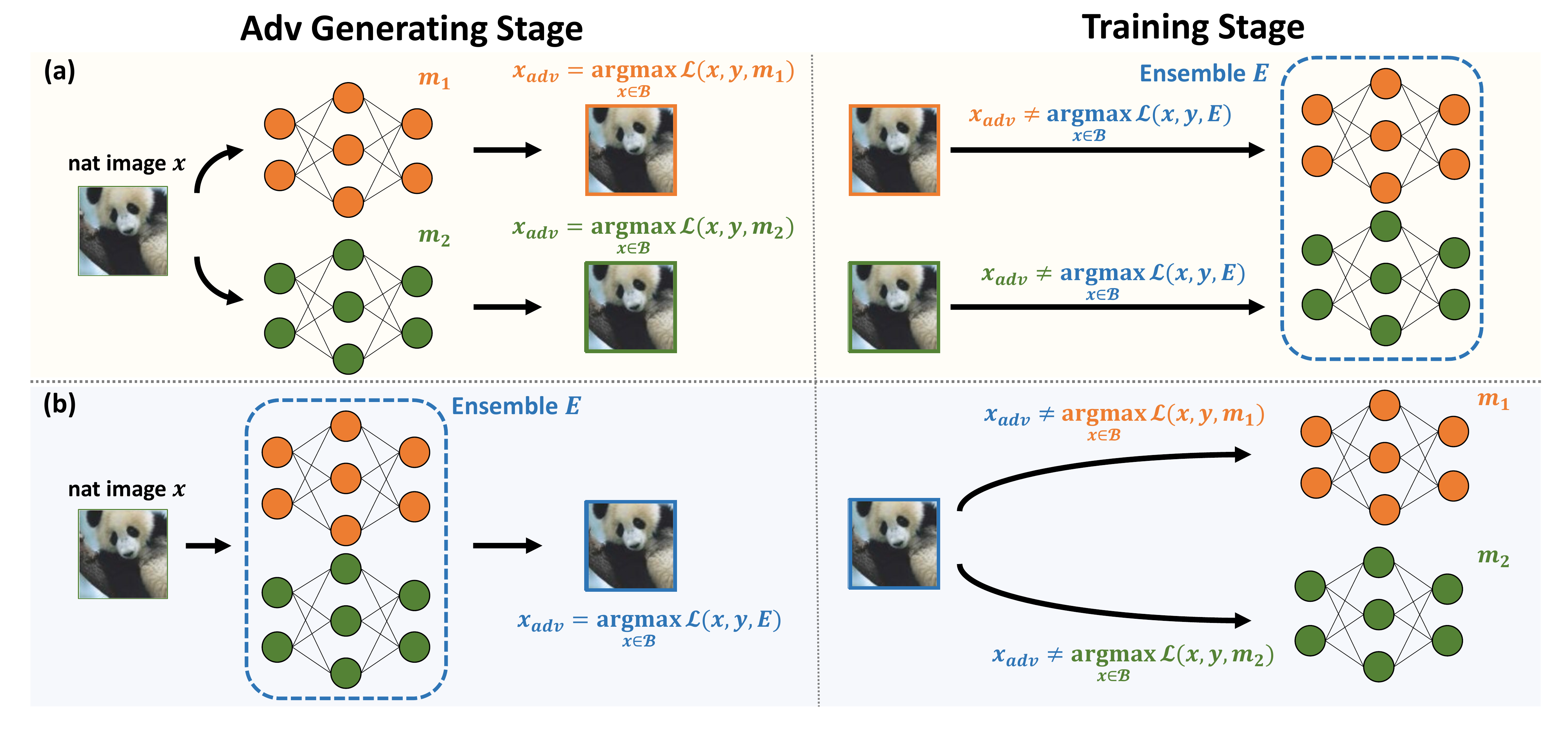} 
  }
  \caption{\textbf{Dilemma of decoupling.} Adversarial training with naive ensemble faces dilemma on choosing target model for generating adversarial examples.  (a)  Adversarial examples generated with respect to each member are suboptimal attack to whole ensemble and less useful to minimize empirical risk. (b)  Adversarial examples generated with respect to whole ensemble are suboptimal to members and make them weak and less diverse, which in consequence loses benefit of ensemble.  
  }
  \label{fig:dod}
\end{figure*}
 
 However, adopting naive ensemble method faces two problems. 
 First and trivial problem is computation cost increments. 
 Adversarial training is already expensive due to multiple gradient computation required in generating adversarial examples. 
 Training multiple networks to ensemble with adversarial examples will multiply the expense accordingly and the overall training becomes overburden.
 
 The other problem not yet discussed much is dilemma on choosing models to generate adversarial examples with respect to. 
 The adversarial characteristic of an example is defined by the corresponding network model.
 In naive ensemble method, there can be two options on which model the adversarial example should be generated with respect to; individual members or whole ensemble system. 
 If adversarial examples are generated with respect to each member, the whole system is trained with suboptimal attack. 
 If adversarial examples are generated with respect to the whole ensemble system, the members of ensembles are trained weakly and are likely to become less diverse. 
 We illustrate the idea in figure~\ref{fig:dod} and name it as dilemma of decoupling. 
 Further elaboration is dealt in proposed method section in detail.

The main purpose of this paper is to introduce the ensemble method which can improve the adversarial training further.
In this paper, we propose stochastic weight averaging method(SWA)\cite{izmailov2018averaging} based adversarial training(SWAAT) that defense white-box attack effectively. 
During adversarial training, weight states on training trajectory are aggregated to the average and the aggregated weight state is updated periodically. 
This makes it possible to have an ensemble effect while it does not suffer from the downside of ensemble we discussed previously.
The adversarial examples generated during the proposed method is optimal to the members and nearly optimal to overall ensemble.
This is a special benefit of adopting SWA in adversarial training which does not belong to SWA in standard training.
Moreover, due to the original characteristic of SWA, additional computation cost can be remained negligible. 
As a result, SWA enables one to train models with stronger robustness efficiently.
  
 The contributions of this paper we claim are as follows:

\begin{itemize}
\item We point out a dilemma of decoupling that arises when using ensemble method in adversarial training.
The dilemma is concerning about which model to choose when generating adversarial examples to train.
\item We propose a SWAAT without the drawbacks of conventional ensembling methods. The proposed method provides not only effective ensembling strategy, but also fast batch normalization method in weight averaging process of SWA and diversification of member models by member-specific data selection stage.
\item Extensive experiments which are conducted using CIFAR-10 and CIFAR-100 datasets show that the proposed method significantly improves performance in defending designed adversarial attacks.
\end{itemize}

\section{Related work}

\subsection{Adversarial training and overfitting}
Schmidt~\cite{schmidt2018adversarially} find that the model shows large performance gap between trained data and test data.
By constructing toy examples, Schmidt provides some theoretical insight that adversarial training requires more complex training samples than standard training. 
Many research have been performed to utilize additional data
~\cite{hendrycks2019using,carmon2019unlabeled,uesato2019labels,zhai2019adversarially} or external regularizers that can mitigate overfitting effects~\cite{zhang2016understanding,zhang2018mixup,yun2019cutmix}.

Recently, there have been works that focus more on the temporal aspect of overfitting in the context of adversarial training. 
\cite{wong2019fast} show that adversarial training with single step attack FGSM~\cite{goodfellow2014explaining} can produce robust model if appropriate noise are induced, and early stopping is performed; this contradict the common belief in adversarial training.
It is analyzed that the adversarial training with FGSM suddenly fails at a moment in the training when it experience catastrophic overfitting phenomena.
\cite{rice2020overfitting} also claim that as the robustness of adversarial trained model degrade as training proceed due to overfitting, there are huge performance gap between best checkpoint aquired during training, and the model at the end of the training. 
According to Rice, simple early stopping can improve adversarial training more than most of external regularizers.
These works provide an insight that the overfitting can be mitigated by focusing more on temporal aspect. 
Therefore, we design SWAAT that updates models to ensembled one frequently during training process.
 
\subsection{Ensemble in Adversarial training}
The most impactful research that connects ensemble and adversarial robustness is the research of Tramer~\cite{tramer2017ensemble}. 
Tramer found that adversarial training with single step attack make the loss surface around data points hard to linearize which slow down adversarial training.
To solve the problem, Tramer trained model with transfered attack from other models which does not have such artifacts. 
This attempt turns out to be successful to improve robustness on black-box condition, but not on white-box one.

Though many researches that connect ensemble and adversarial robustness have proposed, only few surpass adversarial training~\cite{madry2017towards} in white-box condition.
Majority of these researches  more focus on transferability~\cite{kariyappa2019improving,truex2019effects,mahfuz2020ensemble,chow2019denoising}, ensuring theoretic certifications~\cite{lecuyer2019certified,cohen2019certified} or finding alternative solution beside PGD-AT~\cite{pang2019improving,liu2018towards,abbasi2020toward,strauss2017ensemble}.

To the best of our knowledge, there are only two methods related to ensemble that test their model on the strong attack proposed by ~\cite{madry2017towards}, and show better results~\cite{grefenstette2018strength,wang2019resnets}. Although their theoretical motivations are different, These two methods share one common feature; they train the members of ensemble jointly with the adversarial examples generated with respect to the whole ensemble model. In other word, they train the ensemble model like single huge network.

\subsection{Fast Geometric Ensembling and Stochastic weight average}
 
 Garipov~\cite{garipov2018loss} found that there are many points around the local optima that produce similar accuracy but output meaningfully different predictions.
 This suggests that optimizer can found diverse network realizations in the training trajectory. 
 Integrating the idea, Garipov suggested fast geometric ensembling~(FGE) that saves checkpoints from different moments of training and forms the ensemble using it at the inference stage. 
 In other word, the temporal checkpoints during training becomes the members of final ensemble.
 The main advantage of FGE is that almost no computational burden is required to train multiple member models to construct the final ensemble model.
 
 Stochastic weight average~(SWA) is proposed by Izmailov~\cite{izmailov2018averaging} in the literature of work to find better local optima in the loss surface of Deep Neural Network~(DNN) that generalizes better. 
 Although Izmailov gives three explanations that why SWA can be beneficial, here we focus on the behavior of SWA in ensemble perspective. 
 Similar to FGE, SWA also collect checkpoints during training, but instead of forming ensemble with multiple models, it averages the weight states of the checkpoints and update the single model with the averaged weight at the end of training.
 Izmailov shows that a model with averaged weights can approximate averaging multiple models with the weights; i.e. SWA is approximation of FGE.
 
 The strength of SWA compared to FGE is that ensemble model trained with SWA composed of single networks in inference stage also, which result in the computational cost to be remain similar to single model though it has performance of ensemble.
 Moreover, we find that the trait that architecture of members and ensembles being same can be extra useful in adversarial training context as training model can be easily swapped to ensemble model during training which enables one to avoid dilemma of decoupling.

\newcommand{\argmaxF}{\mathop{\mathrm{argmax}}\limits}
\newcommand{\set}[2]{\left\{\  #1  \ \left| \ #2 \ \right. \right\}  }
\section{Propose method}

In this section, we depict the problems of naive ensemble in the context of adversarial training, and propose stochastic weight average based adversarial training~(SWAAT) that does not suffer from the problems.  
We start our discussion by representing the adversarial training in general into two stages : adversarial example generating stage and training stage.  Each stage can be represented as

\begin{equation}
\label{eqn:1}
x_{adv} = \argmaxF_{x'\in \mathcal{B}(x)} L(m_{i}(\theta_{i}, x'), y)
\end{equation}
\begin{equation}
\label{eqn:2}
\theta = \theta_{j} -  \eta\nabla_{\theta}L( m_{j}(\theta_{j}, x_{adv}), y),
\end{equation}

\noindent where $L$ is a loss function, $\eta$ is learning rate of weight parameters, ${\{x,y\}}$ are training data and label pairs, and $\mathcal{B}(x)$ is the area around the original data that constrain the degree of perturbation. 
Commonly, $\mathcal{B}(x)=\set{x'}{\| x-x'\|_\infty < \epsilon }$ or $\mathcal{B}(x)=\set{x'}{\| x-x'\|_2 < \epsilon }$ is selected as the attack constraint. $m(\theta, \cdot)$ is the function of a model with weight states $\theta$. 
If $m_i$, $\theta_i$ in Equation~(\ref{eqn:1}) and $m_j$, $\theta_j$ in Equation~ (\ref{eqn:2}) are same, we call the adversarial example are coupled or optimized to the objective model $m_j(\theta_j, \cdot)$; otherwise, we call the attack is decoupled or suboptimal. For abbreviation, we abuse the notation that the coupled adversarial example $x_{adv}$ to a model function $m(\theta, \cdot)$ as 
\begin{equation}
\label{eqn:3}
x_{adv}(m(\theta), x) = \argmaxF_{x'\in \mathcal{B}(x)} L(m(\theta, x'), y).
\end{equation}

 Also we denote the whole ensemble as $E$ that is composed of $l$ members $m_i$, $i \in [0,l)$ as 
\begin{equation}
\label{eqn:4}
E(x) = \bigcup_{i=0}^{l-1}m_i(\theta_i, x),
\end{equation}

\noindent and alternatively, $ m_i(\theta_i) \subset E$.
The naive ensemble method we refer is the late ensemble which uses the results of member models which can be represented as
\begin{equation}
\label{eqn:5}
E(x) = \frac{1}{l}\sum_{i=0}^{l-1}m_i(\theta_i, x),
\end{equation}

\noindent while the summation can be substituted by majority voting.
 In naive ensemble training, the models can be trained with examples adversarial examples optimized to a member of ensemble: $x_{adv}(m_i(\theta), x)$.
Otherwise, it can also be optimized to the whole ensemble: $x_{adv}(E, x)$

\begin{figure*}[t]
\centering
\scalebox{0.9}{
  \includegraphics[width=0.95\textwidth]{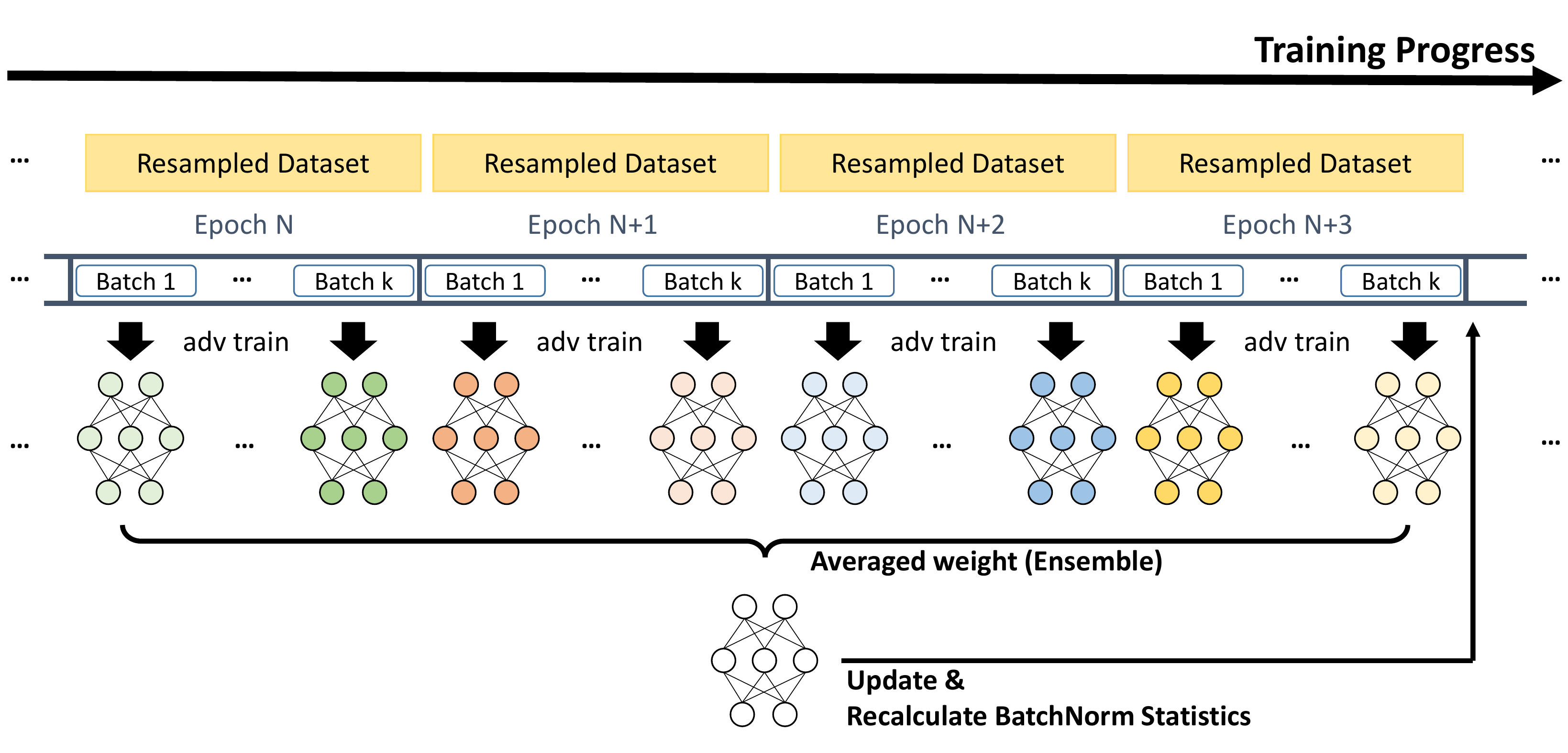} 
  }
  \caption{\textbf{Stochastic Weight Average based Adversarial Training~(SWAAT.)} SWAAT aggregates the temporal weight states during training and replaces the model weight with the aggregated one at the end of each epoch. SWAAT-window is 4 in this figure.
}
\label{fig:swaat}
\end{figure*}

\subsection{Dillema of Decoupling}
There is a dilemma on choosing target model in adversarial example generation stage as shown in Figure~\ref{fig:dod}.
 Tramer~\cite{tramer2017ensemble} found that adversarial training with single-step methods admits a degenerate global minimum problem which result in the attack methods do not provide perturbation that fools the model. 
In the paper, to solve the problem, decoupled adversarial examples are used for adversarial training in addition to coupled examples. 
 Although the method was successful to improve robustness under black-box attack scenario, it showed worst robustness in white-box scenario compared to the model trained with coupled examples only.
 According to analysis of Tramer, that this is because the model is trained with less coupled adversarial examples.
 Therefore, it can be conjectured that coupling the attack to the objective model is crucial to secure robustness in white-box attack scenario. 

In this perspective, it is significant to train the models with $x_{adv}(E, x)$ instead of $x_{adv}(m_i, x)$.
As ensemble $E$ is the final model that is used in inference, training examples need to be coupled with $E$; otherwise, the overall system $E$ will be vulnerable in white-box scenario.
Result of Grefenstette et al.~\cite{grefenstette2018strength} also support the ensemble schemes that trains $x_{adv}(E,x)$.
In the paper, ensembling the models $m_i(\theta_i, x) \subset E$ separately trained with their own adversarial $x_{adv}(m_i(\theta_i),x)$ showed worse robustness compared to the ensemble model trained with the attacks $x_{adv}(E,x)$ that are coupled to the whole ensemble.

However, training models with $x_{adv}(E,x)$ loses the benefit of ensemble.
When the number of members is restricted, the performance of overall ensemble is closely related to the performance of each member and diversity of members. 
However, $x_{adv}(E,x)$ is decoupled to each member $m_i$ which will consequently make the members to have weak robustness.
Moreover, training the members with same adversarial example $x_{adv}(E,x)$ will degenerates the diversity of members.
In case of training all member models with same adversarial example, the members come to share the adversarial subspace, or vulnerability, and chance to repulse adversarial attempts from disagreement will be reduced.
Therefore, the benefits from ensemble will diminish and ensemble model will act like single huge model. 
We call this dilemma in naive ensemble method as dilemma of decoupling.

\subsection{SWA based Adversarial Training}
\textbf{Adoption of SWA.} In this section, we propose SWA based adversarial training~(SWAAT).
The main characterestic of SWA is that it utilize single network with averaged weight instead of averaging multiple networks. 
The weight to be averaged is collected and aggregated during training.
 In original SWA, temporal weight states at the \textit{end of each epoch} are aggregated to buffer by cumulative moving average, and model is updated with the aggregated weight \textit{at the end of the training}.

 In our SWAAT, on the other hand, the temporal weight states \textit{after every training iteration} are aggregated to buffer by simple moving average, and model is updated with the aggregated weight \textit{at the end of each epoch}.
 For abbreviation, we denote SWAAT has SWAAT-window with size $M$ epochs if SWAAT aggregates $M \cdotp k$ temporal weight states where $k$-iteration completes an epoch of training.
 Moreover, if $n$-iteration completes $N$ epochs of training, SWAAT model with SWAAT-window $M$ at the end of $N$-th epoch can be represented
 \begin{equation}
\label{eqn:5}
E_N(x) = m(\theta_{swa},x) = m(\frac{1}{Mk}\sum_{i \in S}\theta^i, x), 
\end{equation}

\noindent while $\theta^i$ is the weight state of a model $m$ after $i$-th iteration in training, and $S$ is iteration set in SWAAT-window, i.e., $S=\set{i}{n-M k +1 < i < n }$.

 Izmailov~\cite{izmailov2018averaging} claimed that if weight states are close to each other, a model with averaged weight approximates averaging model with the weight, i.e,
  \begin{equation}
\label{eqn:6}
 m(\frac{1}{M  k}\sum_{i \in S}\theta_i, x)   \approx \frac{1}{M  k} \sum_{i \in S}m(\theta_i, x)  
\end{equation}
 and therefore, it can be described that
 
  \begin{equation}
\label{eqn:7}
m(\theta_{swa},x) = \bigcup_{i \in S}m(\theta^i, x)  
\end{equation}
 
 We note that one important difference in SWAAT from SWA is that weight of the model is updated to aggregated weight $\theta_{swa}$ at the end of every epoch instead of final epoch.

While SWA are effectively similar to ensemble model, the computational cost of it is comparable to that of the SGD counterpart. 
The only additional operation is weight state aggregation which can be implemented efficiently by using moving average method and recalculating statistic related to batch normalization at the end of the training.
Basically, SWAAT follows the mechanism of SWA and therefore it reduces generalization gap by taking advantage of ensemble effect without significant computation cost increase. 
 
Moreover, the dilemma of decoupling is not problematic in SWAAT.
As we update the model to ensemble model at the end of every epoch, the initial point of the model at epoch N is an ensembled model  $E_{N-1}$.
In other words, training process in $N$-th epochs can be regarded as adversarial training the ensemble model $E_{N-1}$.
We remark that in normal SGD-based adversarial training, the adversarial minibatch $x_{adv}(m(\theta_i), x)$ that is trained in $i$-th iteration is slightly decoupled to the model at different point in training, $m(\theta_{j}, x)|_{j \ne i}$, as weight parameters are updated.
In SWAAT, ensemble model is decoupled only as much as the model is in normal SGD based adversarial training.
It can be more easily understood when one omits the aggregation at the end of training.
Also, it is obvious that each minibatch of adversarial examples is coupled to the corresponding temporal state of the model which is a member of ensemble.
Therefore, the adversarial examples trained during SWAAT is coupled with the members and not decoupled severely with the whole ensemble and does not suffer from decoupling dilemma. \\

\begin{algorithm}[t]
\SetAlgoLined
\DontPrintSemicolon
    \SetKwInOut{Input}{Input}
    \SetKwInOut{Output}{Output}
    
    \KwSty{Constant :}   SWAAT-window $M$, \; total training epoch $N_{total}$, iterations per epoch $k$\;
    \Input{model $m(\theta^{0})$, training dataset $\mathcal{D}_{whole}$\;}
    \Output{robust model $m(\theta_{swa})$}
    \KwSty{Initialize  }   
    $\mathcal{D} \leftarrow \mathcal{D}_{whole} $,
    $\theta_{swa} \leftarrow 0$\;
    \For{\ProcNameSty{iteration} i = 0,1, ... N\textsubscript{total}}
      {
        $x \leftarrow$ \ProcNameSty{Get$\_$batch}$(\mathcal{D})$ \;
        $x_{adv} \leftarrow$ \ProcNameSty{PGD}$(m(\theta^i),x)$ \;
        $\theta^{i+1} \leftarrow$ \ProcNameSty{SGD}$(m(\theta^i),x_{adv})$ \;
        $w \leftarrow \min(i, M k)$ \;
        $\theta_{swa}\leftarrow \frac{(w-1)}{w}\theta_{swa}+\frac{1}{w}\theta^{i+1}$
        
        \If{\ProcNameSty{mod}(i,k) = 0}
        {
        
        $\theta^{i+1} \leftarrow \theta^{swa}$\;
        $E \leftarrow m(\theta^{i+1})$
        
        $E \leftarrow$  \ProcNameSty{Adjust$\_$BN}$(E,\mathcal{D}_{whole} )$\;
        $\mathcal{D} \leftarrow$ 
        \ProcNameSty{HEM}
        $(E,\mathcal{D}_{whole})$ \;
        $\mathcal{D} \leftarrow $
        \ProcNameSty{Sample$\_$with$\_$replacement}
        $(\mathcal{D})$
        }
      }
    \caption{Stochastic Weight Average based  \;  Adversarial Training}
    \label{alg:SWAAT}
    \vspace{-0.1cm}
\end{algorithm}

\noindent \textbf{Batch Normalization Statistic.}
When weight states are averaged in SWA, mean and standard deviation of hidden response in network should be recalculated respectively. Additional cost for this recalculation in SWA is neglectable as the model is updated to $\theta_{swa}$ once.
On the other hand, in the proposed SWAAT, weights of the model are aggregated and updated for every epoch.
Also, adversarial examples are generated in evaluation mode which uses the statistic of the whole data.
Therefore, to generate adversarial example properly, the batch normalization statistic of adversarial example should be recalculated after every aggregation. 
The recalculation incurs overheads as it additionally requires generating adversarial examples for every epoch. 

Fortunately, we empirically found that recalculating the statistics with natural image gives similar performance at the end to that with adversarial examples.
Therefore, it is unnecessary to generate adversarial examples additionally.
This is surprising because adversarial examples are expected to have significantly different statistic on response from that of normal examples.
More studies are required on this phenomenon.\\

\noindent \textbf{Diversity via Data Resampling.}
 SWAAT can be improved by introducing more diversity to the members. 
 In original SWA, Izmailov used cyclic learning rate which can find more diverse points. 
 However, the attempt was not successful; it was empirically shown that models trained with such cyclic learning rate perform worse than constant learning rate counterpart.
 
 We propose another way that delivers diversity to the members of SWAAT.
 Adversarial training can be regarded as very sensitive to the training data because it overfits easily.
This means that slight change in training dataset can introduce diversity on members.
Hence, we resample the data from the original dataset with replacement while fixing the size of resampled one to be same with the original dataset.
 
 Three resampling strategies are tested in this paper. 
 The first one is resampling with bootstrapped~(BOOT). 
 Each example in the original dataset has same probability to be sampled and the sampling is performed at the beginning of every epoch. 
 As BOOT is just a random sampling, computational overhead of negligible.
 
 The second one is resampling with hard example mining~(HEM). 
 When the weights are aggregated and Ensemble $E_N$ has been formed, we evaluate the model with the training set, and mark the examples that are misclassified by $E$.
 At the end of each epoch, training data for the next epoch is sampled while the misclassified examples have tripled probability to be sampled compared to the correctly classified ones. 
 However, hard example mining can be costly as it requires additional adversarial example generation. To deal with the problem, we tried HEM version with less frequent mining also.
 
 The last one is resampling with online-hard example mining~(OHEM). 
 During the adversarial training in each epoch, the misclassified training examples are marked.
 Although this method does not select hard examples of $E$ exactly, it does not require additional adversarial example generations.
 Therefore, additional computational cost is negligible in this scheme.
 We outline an example of our method with pseudo code Algorithm~\ref{alg:SWAAT}. In the example, PGD is used for adversarial training and data resampling is done with hard example mining at the end of every epoch.


\noindent
\begin{table*}[t]

\begin{center}
\begin{tabularx}{0.90\textwidth}{c|c|c|c|c|c|c}
\toprule
\multicolumn{1}{c|}{\textbf{Dataset}} & \textbf{Method}   & \textbf{Architecture} & \textbf{Natural} & \textbf{PGD-10} & \textbf{PGD-20} & \textbf{CW} ($l_2$)  \\ \hline
\multirow{12}{*}{CIFAR10}    & PGD-AT~\shortcite{madry2017towards}\textsuperscript{*} &  preact-18   &  82.91(82.98)  & 53.01(52.69)  &   52.76(52.54)   &  78.10 \\ 
& PGD-AT~\shortcite{madry2017towards}\textsuperscript{*}   &  wrn-28-10       &  87.01(87.08)       &   56.44(56.17) &  56.23(56.02)        &     79.24    \\
    & TRADE~\shortcite{zhang2019theoretically}\textsuperscript{*}    &  preact-18       &    81.94(82.11)     &     51.35(50.99)   & 51.01(50.74)       &     79.18       \\
    & TRADE~\shortcite{zhang2019theoretically}    &  wrn-34-10       &    84.92     &      56.6   &   56.43  &   -             \\
    & Pretrain\shortcite{hendrycks2019using}      & wrn-34-10     &      87.1-   &     -   &    57.4-  &    - \\
    & LLR~\shortcite{qin2019adversarial}    &  wrn-28-8       &    86.83     &     -   &   52.99     &     -                                 \\
    & ATES~\shortcite{sitawarin2020improving}    &  wrn-34-10       &    86.84    &     -   &   53.18     &     -                                \\
    & RSE\shortcite{liu2018towards}\textsuperscript{$\dagger$}      &  VGG-16        &    -     &   -     &      -  &     34.75 \\
    & Naive~\shortcite{grefenstette2018strength}\textsuperscript{$\dagger$}    &   wrn-28-10 $\times$4      &  88.-       &    -    &      52.-  &     -                                \\
    & EnRes\shortcite{wang2019resnets}\textsuperscript{$\dagger$}    &    wrn-34-10 &   86.19      &   -     &  56.60            &     -               \\
    & SWAAT(Ours)\textsuperscript{*}    & preact-18        &   83.85(83.89)      &   58.56(58.32)     & 58.23(58.11)       &    81.66                                  \\
    & SWAAT(Ours)\textsuperscript{*}    &  wrn-28-10       &   \textbf{87.98(87.93)}      &    \textbf{60.94(60.69)}    &  \textbf{60.57(60.33)}   &        \textbf{83.04}             \\ \hline
\multirow{7}{*}{CIFAR100} & PGD-AT~\shortcite{madry2017towards}\textsuperscript{*}   & preact-18        &     51.17 (51.31)   & 25.95(25.71)       &  25.64(25.52)      &           48.97                             \\
    & PGD-AT~\shortcite{madry2017towards}\textsuperscript{*}   & wrn-28-10        &  57.77(58.19)       &     28.75(28.53)   &    28.62(28.34)    &                         49.11           \\
     & TRADE~\shortcite{zhang2019theoretically}\textsuperscript{*}    & preact-18        &   53.07(52.90)      & 25.04(24.78)       &    24.82(24.66)    &                    -                  \\
    & Pretrain\shortcite{hendrycks2019using}      & preact18        &     59.2    &     -   &           \textbf{33.5-}     &         -                       \\
    & EnRes~\shortcite{wang2019resnets}\textsuperscript{$\dagger$}    &  preact-20 $\times$5      &      51.72   &     -   &       27.80  &             -                        \\
    & SWAAT(Ours)\textsuperscript{*}\textsuperscript{$\dagger$}    & preact-18         & 55.08(55.01)  &     28.61(28.43)   &  28.27(28.19)      &     49.20                             \\
    & SWAAT(Ours)\textsuperscript{*}\textsuperscript{$\dagger$}    &  wrn-28-10  &    \textbf{59.84(59.67)}     &     \textbf{31.84(31.81)}   &  31.07(30.93)      &  \textbf{49.83}                                    \\ \hline               
\multirow{2}{*}{SVHN}                          & PGD-AT~\shortcite{madry2017towards}\textsuperscript{*}  &  preact-18   &     91.76(90.49)    &   67.35(57.40)     & 58.50(55.87)       &   23.50                                   \\
& SWAAT(Ours)\textsuperscript{*}    &  preact-18       &  \textbf{92.61(92.63)}  & \textbf{62.98(62.57)}  &   \textbf{61.48(61.21)}     & \textbf{60.37}   \\ 
\bottomrule
\end{tabularx}
\end{center}
\caption{\textbf{Adversarial accuracy on CIFAR-10, CIFAR-100, and SVHN  datasets.} $*$ indicates that the result are from our own implementation; otherwise the result from the official papers are reported. $\dagger$ indicates the method uses ensemble concepts. Numbers in parentheses indicates the average of 5 adjacent epochs}
\label{tb1}
\end{table*}

\section{Experimental Results}

\subsection{Implementation Detail}
The proposed idea is implemented by training three popular benchmark datasets: CIFAR10, CIFAR-100, and SVHN.
We use pre-activation ResNet18~(preact-18)~\cite{he2016identity} as our default model to train and trained WideResNet-28-10~(wrn-28-10)~\cite{Zagoruyko2016WRN} additionally for CIFAR-10 and CIFAR-100. 

To set the baseline, we reimplement projected gradient descent based adversarial training ~(PGD-AT), following the original training protocol~\cite{madry2017towards} mostly.
The only difference from the original protocol is that we train models with adversarial examples generated by 10 step iterations instead of $7$.

SWAAT uses similar protocols to the baseline model, and therefore, we only describe the difference.
For CIFAR-10 and CIFAR-100, checkpoint at 100-th epoch of the baseline models are used as the initial point of SWAAT. 
This is similar to that original SWA uses model trained with conventional SGD as its initial point.
On the other hand, in case of SVHN, we find that the moment model starts to overfit is very inconsistent and use SWAAT from scratch. 
For fair comparison, we train SWAAT for 100 epochs on CIFAR-10 and CIFAR-100 to fix the total training length to 200 epochs.

Regardless the datasets, SWAAT-window is set to four. For example, as one epoch of CIFAR-10 is composed of 391 iterations, SWAAT aggregates weight states of $4\times391$ latest iterations.
Similar to the original SWA that uses higher learning rate compared to that of conventional SGD training, we modify the learning rate of SWAAT to be higher than that of the baseline.
However, different from SWA, we modify the learning rate adaptively to SWAAT-window when we try multiple configurations of SWAAT-window. 
We set the learning rate of SWAAT to be SWAAT-window times larger than the learning rate of the baseline at every moment of training; i.e., as the learning rate in $100$-th epoch in CIFAR-10 training is 0.01, the learning rate of SWAAT starts from 0.04.

The effect of individual implementation elements are described in supplementary materials 

\subsection{Main results}
As our proposed method targets the white-box scenario, we evaluate our model under PGD attack~\cite{madry2017towards} and Carlini-Wanger attack~(CW)~\cite{carlini2017towards}. 
PGD is considered as a universal “first-order adversary”, i.e., the strongest attack utilizing the local first order information about the network.
Under $l_{\infty}$ constraint, PGD with step-size $\alpha = 2/255$ and radius $\epsilon = 8/255$ was used as the attack following the original paper~\cite{madry2017towards}.
We notate $n$-step PGD as PGD-$n$. PGD-10 and PGD-20 are tested as the benchmark.
Under $l_{2}$ constraint, we test CW attack with $c=0.2$, $\kappa=0$, and learning rate 0.01. This is to compare with \cite{liu2018towards} mainly.

The performance that we report is that of the best performing checkpoint we get during training to not get penalty from the protocol.
We remark that Rice claim PGD-AT with early stopping outperforms TRADES~\cite{zhang2019theoretically} with early stopping which is considered as state-of-the-arts; we observe similar results. 
In short, PGD-AT with early stopping is a strong baseline.
To prevent the unstable training methods we also report the average of accuracy on fiv epochs to the best ones.

We show the performance of our models on Table~\ref{tb1}. 
The result of SWAAT with HEM that mines hard example for every epoch is reported only, as it shows the highest accuracy.
Beside of reporting the result of our own experiments, we represent the results from other recent defense methods in the table to show how robust the SWAAT model is.

To the best of our knowledge, the only work that takes benefit of ensemble and surpasses PGD-AT under white-box PGD attack is the method introduced by ~\cite{wang2019resnets}, and ~\cite{grefenstette2018strength}.
Beside two method, ~\cite{liu2018towards} also claim that their method is comparable in optimization based attack CW.

As one can observe, a model trained with SWAAT exceed other methods in all most all experiments with even if it has smaller architecture such as preact-18.
We also remark that pretrain method~\cite{hendrycks2019using} utilizes downsampled Imagenet data~\cite{chrabaszcz2017downsampled} for pretraining.
In conclusion, even though SWAAT is simple modification of PGD-AT, it improve the robustness of models significantly.

\subsection{Comparison of Data Resampling Strategy}
In this subsection, we compare the influence of different data resampling methods discussed previously. 
For HEM, we try several versions that mines hard examples with different frequently. 
HEM-$N$ means that hard examples are mined for every $N$ epochs while resampling is performed every epoch.
We found that BOOT only makes the robustness of models worse. 
It can be analyzed that the negative effect from not seeing whole data is bigger than the positive effect from bringing diversity to the members.
The results from different versions of HEM give clear message; performing hard example mining more result in better robustness. 
However, hard example mining requires additional adversarial example generation, and therefore, there is tradeoff between adversarial accuracy and computational cost.
The model trained with OHEM shows reasonable performance considering that OHEM does not cause severe computational cost increase.

\begin{table}[t]
\scalebox{0.95}{
\begin{tabular}{l|l}
\toprule
\multicolumn{1}{c|}{\textbf{Data resampling strategy}}       & \textbf{Accuracy} \\ \hline
without resampling (baseline)                     & 56.08  (+0.00)           \\ \hline
Boot strapping (BOOT)                   & 55.67     (-0.41)        \\ \hline
Hard example mining / 1 epoch (HEM-1) & 58.56     (+2.48)        \\ \hline
Hard example mining / 2 epoch (HEM-2) & 58.13     (+2.05)        \\ \hline
Hard example mining / 4 epoch (HEM-4) & 57.49     (+1.41)        \\ \hline
Online hard example mining (OHEM)       & 57.04     (+0.98)        \\ 
\bottomrule
\end{tabular}
}
\caption{\textbf{Comparison on data resampling methods.} PGD-10, preact-18, CIFAR-10 is used. Numbers in parentheses indicates the improvement from the baseline}
\end{table}

\subsection{Obfuscated Gradient Test}
The validity of defense methods has arisen to a serious issue in the research field. 
To reveal whether our defense relies on obfuscated gradient or not, we conduct two experiments that are introduced by Athalye~\cite{athalye2018obfuscated}.
We evaluate our model with 1000-step PGD~(PGD-1k) with $l_{\infty}$ bound $\epsilon=8/255$ which is expected to show similar accuracy and PGD-100 with $\epsilon=\infty$ which is expected to show accuracy.
We display the result of experiments on Table~\ref{tab:obfg}.
The performance of our models merely degrade on PGD-1k compared to PGD-10, and become zero on unconstrained attack. This indicate our defense is not result from obfuscated gradient.

\begin{table}[t]
\scalebox{0.95}{
\begin{tabular}{c|c|c|c|c}
\toprule
\textbf{Dataset}          & \multicolumn{1}{c|}{\textbf{Model}} & \textbf{PGD-10} & \textbf{PGD-1k} & \textbf{$\epsilon = \infty$} \\ \hline
\multirow{2}{*}{CIFAR10}  & preact-18                           &         58.56        &    58.13               & 0.00                     \\ \cline{2-5} 
                          & wrn-28-10                              & 60.69           &      59.72             & 0.00                     \\ \hline
\multirow{2}{*}{CIFAR100} & preact-18                           & 28.53           &          28.21         & 0.00                     \\ \cline{2-5} 
                          & wrn-28-10                              & 31.84           &    31.26               & 0.00                     \\ \hline
SVHN                      & preact-18                           &   62.98              &    60.42               & 0.00                    \\ 
\bottomrule
\end{tabular}
}
\label{tab:obfg}
\caption{\textbf{Obfuscated gradient Test.} 10 steps, 1000 steps PGD with $\epsilon = 8/255$  and 100 step PGD with $\epsilon = \infty$ has been tested. The result shows that our method is not a result of obfuscated gradients} 
\label{tab:obfg}
\end{table}

\section{Conclusion}

Adversarial attack is a serious issue in modern computer vision society. 
The robustness that can be acquired from adversarial training is limited due to its susceptibility to overfitting. 
To break the limitation, we proposed SWAAT that take benefit from ensemble without severe computation cost increment. 
The experiments shows that SWAAT is the most effective ensemble method in the context of adversarial robustness. 
This is because the diversity and robustness of individual members are considered in training stage. 
We believe that our study brings new insight to the field and accelerate the the adoption of machine learning technologies to real world.

\bibliography{ref}

\begin{thebibliography}{39}
\providecommand{\natexlab}[1]{#1}
\providecommand{\url}[1]{\texttt{#1}}
\providecommand{\urlprefix}{URL }
\expandafter\ifx\csname urlstyle\endcsname\relax
  \providecommand{\doi}[1]{doi:\discretionary{}{}{}#1}\else
  \providecommand{\doi}{doi:\discretionary{}{}{}\begingroup
  \urlstyle{rm}\Url}\fi

\bibitem[{Abbasi et~al.(2020)Abbasi, Rajabi, Gagn{\'e}, and
  Bobba}]{abbasi2020toward}
Abbasi, M.; Rajabi, A.; Gagn{\'e}, C.; and Bobba, R.~B. 2020.
\newblock Toward adversarial robustness by diversity in an ensemble of
  specialized deep neural networks.
\newblock In \emph{Canadian Conference on Artificial Intelligence}, 1--14.
  Springer.

\bibitem[{Athalye, Carlini, and Wagner(2018)}]{athalye2018obfuscated}
Athalye, A.; Carlini, N.; and Wagner, D. 2018.
\newblock Obfuscated gradients give a false sense of security: Circumventing
  defenses to adversarial examples.
\newblock \emph{arXiv preprint arXiv:1802.00420} .

\bibitem[{Carlini and Wagner(2017)}]{carlini2017towards}
Carlini, N.; and Wagner, D. 2017.
\newblock Towards evaluating the robustness of neural networks.
\newblock In \emph{2017 ieee symposium on security and privacy (sp)}, 39--57.
  IEEE.

\bibitem[{Carmon et~al.(2019)Carmon, Raghunathan, Schmidt, Duchi, and
  Liang}]{carmon2019unlabeled}
Carmon, Y.; Raghunathan, A.; Schmidt, L.; Duchi, J.~C.; and Liang, P.~S. 2019.
\newblock Unlabeled data improves adversarial robustness.
\newblock In \emph{Advances in Neural Information Processing Systems},
  11192--11203.

\bibitem[{Chow et~al.(2019)Chow, Wei, Wu, and Liu}]{chow2019denoising}
Chow, K.-H.; Wei, W.; Wu, Y.; and Liu, L. 2019.
\newblock Denoising and Verification Cross-Layer Ensemble Against Black-box
  Adversarial Attacks.
\newblock In \emph{2019 IEEE International Conference on Big Data (Big Data)},
  1282--1291. IEEE.

\bibitem[{Chrabaszcz, Loshchilov, and Hutter(2017)}]{chrabaszcz2017downsampled}
Chrabaszcz, P.; Loshchilov, I.; and Hutter, F. 2017.
\newblock A downsampled variant of imagenet as an alternative to the cifar
  datasets.
\newblock \emph{arXiv preprint arXiv:1707.08819} .

\bibitem[{Cohen, Rosenfeld, and Kolter(2019)}]{cohen2019certified}
Cohen, J.; Rosenfeld, E.; and Kolter, Z. 2019.
\newblock Certified Adversarial Robustness via Randomized Smoothing.
\newblock In \emph{International Conference on Machine Learning}, 1310--1320.

\bibitem[{Dietterich(2000)}]{dietterich2000ensemble}
Dietterich, T.~G. 2000.
\newblock Ensemble methods in machine learning.
\newblock In \emph{International workshop on multiple classifier systems},
  1--15. Springer.

\bibitem[{Garipov et~al.(2018)Garipov, Izmailov, Podoprikhin, Vetrov, and
  Wilson}]{garipov2018loss}
Garipov, T.; Izmailov, P.; Podoprikhin, D.; Vetrov, D.~P.; and Wilson, A.~G.
  2018.
\newblock Loss surfaces, mode connectivity, and fast ensembling of dnns.
\newblock In \emph{Advances in Neural Information Processing Systems},
  8789--8798.

\bibitem[{Goodfellow, Shlens, and Szegedy(2014)}]{goodfellow2014explaining}
Goodfellow, I.~J.; Shlens, J.; and Szegedy, C. 2014.
\newblock Explaining and harnessing adversarial examples.
\newblock \emph{arXiv preprint arXiv:1412.6572} .

\bibitem[{Grefenstette et~al.(2018)Grefenstette, Stanforth, O'Donoghue, Uesato,
  Swirszcz, and Kohli}]{grefenstette2018strength}
Grefenstette, E.; Stanforth, R.; O'Donoghue, B.; Uesato, J.; Swirszcz, G.; and
  Kohli, P. 2018.
\newblock Strength in Numbers: Trading-off Robustness and Computation via
  Adversarially-Trained Ensembles.
\newblock \emph{arXiv preprint arXiv:1811.09300} .

\bibitem[{He et~al.(2016)He, Zhang, Ren, and Sun}]{he2016identity}
He, K.; Zhang, X.; Ren, S.; and Sun, J. 2016.
\newblock Identity mappings in deep residual networks.
\newblock In \emph{European conference on computer vision}, 630--645. Springer.

\bibitem[{Hendrycks, Lee, and Mazeika(2019)}]{hendrycks2019using}
Hendrycks, D.; Lee, K.; and Mazeika, M. 2019.
\newblock Using pre-training can improve model robustness and uncertainty.
\newblock \emph{arXiv preprint arXiv:1901.09960} .

\bibitem[{Izmailov et~al.(2018)Izmailov, Podoprikhin, Garipov, Vetrov, and
  Wilson}]{izmailov2018averaging}
Izmailov, P.; Podoprikhin, D.; Garipov, T.; Vetrov, D.; and Wilson, A.~G. 2018.
\newblock Averaging weights leads to wider optima and better generalization.
\newblock \emph{arXiv preprint arXiv:1803.05407} .

\bibitem[{Kariyappa and Qureshi(2019)}]{kariyappa2019improving}
Kariyappa, S.; and Qureshi, M.~K. 2019.
\newblock Improving adversarial robustness of ensembles with diversity
  training.
\newblock \emph{arXiv preprint arXiv:1901.09981} .

\bibitem[{Lecuyer et~al.(2019)Lecuyer, Atlidakis, Geambasu, Hsu, and
  Jana}]{lecuyer2019certified}
Lecuyer, M.; Atlidakis, V.; Geambasu, R.; Hsu, D.; and Jana, S. 2019.
\newblock Certified robustness to adversarial examples with differential
  privacy.
\newblock In \emph{2019 IEEE Symposium on Security and Privacy (SP)}, 656--672.
  IEEE.

\bibitem[{Liu et~al.(2018)Liu, Cheng, Zhang, and Hsieh}]{liu2018towards}
Liu, X.; Cheng, M.; Zhang, H.; and Hsieh, C.-J. 2018.
\newblock Towards robust neural networks via random self-ensemble.
\newblock In \emph{Proceedings of the European Conference on Computer Vision
  (ECCV)}, 369--385.

\bibitem[{Madry et~al.(2017)Madry, Makelov, Schmidt, Tsipras, and
  Vladu}]{madry2017towards}
Madry, A.; Makelov, A.; Schmidt, L.; Tsipras, D.; and Vladu, A. 2017.
\newblock Towards deep learning models resistant to adversarial attacks.
\newblock \emph{arXiv preprint arXiv:1706.06083} .

\bibitem[{Mahfuz, Sahay, and Gamal(2020)}]{mahfuz2020ensemble}
Mahfuz, R.; Sahay, R.; and Gamal, A.~E. 2020.
\newblock Ensemble Noise Simulation to Handle Uncertainty about Gradient-based
  Adversarial Attacks.
\newblock \emph{arXiv preprint arXiv:2001.09486} .

\bibitem[{Pang et~al.(2019)Pang, Xu, Du, Chen, and Zhu}]{pang2019improving}
Pang, T.; Xu, K.; Du, C.; Chen, N.; and Zhu, J. 2019.
\newblock Improving Adversarial Robustness via Promoting Ensemble Diversity.
\newblock In \emph{International Conference on Machine Learning}, 4970--4979.

\bibitem[{Papernot et~al.(2016)Papernot, McDaniel, Wu, Jha, and
  Swami}]{papernot2016distillation}
Papernot, N.; McDaniel, P.; Wu, X.; Jha, S.; and Swami, A. 2016.
\newblock Distillation as a defense to adversarial perturbations against deep
  neural networks.
\newblock In \emph{2016 IEEE Symposium on Security and Privacy (SP)}, 582--597.
  IEEE.

\bibitem[{Qin et~al.(2019)Qin, Martens, Gowal, Krishnan, Dvijotham, Fawzi, De,
  Stanforth, and Kohli}]{qin2019adversarial}
Qin, C.; Martens, J.; Gowal, S.; Krishnan, D.; Dvijotham, K.; Fawzi, A.; De,
  S.; Stanforth, R.; and Kohli, P. 2019.
\newblock Adversarial robustness through local linearization.
\newblock In \emph{Advances in Neural Information Processing Systems},
  13847--13856.

\bibitem[{Rice, Wong, and Kolter(2020)}]{rice2020overfitting}
Rice, L.; Wong, E.; and Kolter, J.~Z. 2020.
\newblock Overfitting in adversarially robust deep learning.
\newblock \emph{arXiv preprint arXiv:2002.11569} .

\bibitem[{Schmidt et~al.(2018)Schmidt, Santurkar, Tsipras, Talwar, and
  Madry}]{schmidt2018adversarially}
Schmidt, L.; Santurkar, S.; Tsipras, D.; Talwar, K.; and Madry, A. 2018.
\newblock Adversarially robust generalization requires more data.
\newblock In \emph{Advances in Neural Information Processing Systems},
  5014--5026.

\bibitem[{Sitawarin, Chakraborty, and Wagner(2020)}]{sitawarin2020improving}
Sitawarin, C.; Chakraborty, S.; and Wagner, D. 2020.
\newblock Improving Adversarial Robustness Through Progressive Hardening.
\newblock \emph{arXiv preprint arXiv:2003.09347} .

\bibitem[{Song et~al.(2018)Song, Kim, Nowozin, Ermon, and
  Kushman}]{song2018pixeldefend}
Song, Y.; Kim, T.; Nowozin, S.; Ermon, S.; and Kushman, N. 2018.
\newblock PixelDefend: Leveraging Generative Models to Understand and Defend
  against Adversarial Examples.
\newblock In \emph{International Conference on Learning Representations}.

\bibitem[{Strauss et~al.(2017)Strauss, Hanselmann, Junginger, and
  Ulmer}]{strauss2017ensemble}
Strauss, T.; Hanselmann, M.; Junginger, A.; and Ulmer, H. 2017.
\newblock Ensemble methods as a defense to adversarial perturbations against
  deep neural networks.
\newblock \emph{arXiv preprint arXiv:1709.03423} .

\bibitem[{Tram{\`e}r et~al.(2017)Tram{\`e}r, Kurakin, Papernot, Goodfellow,
  Boneh, and McDaniel}]{tramer2017ensemble}
Tram{\`e}r, F.; Kurakin, A.; Papernot, N.; Goodfellow, I.; Boneh, D.; and
  McDaniel, P. 2017.
\newblock Ensemble adversarial training: Attacks and defenses.
\newblock \emph{arXiv preprint arXiv:1705.07204} .

\bibitem[{Truex et~al.(2019)Truex, Liu, Gursoy, Wei, and Yu}]{truex2019effects}
Truex, S.; Liu, L.; Gursoy, M.~E.; Wei, W.; and Yu, L. 2019.
\newblock Effects of differential privacy and data skewness on membership
  inference vulnerability.
\newblock In \emph{2019 First IEEE International Conference on Trust, Privacy
  and Security in Intelligent Systems and Applications (TPS-ISA)}, 82--91.
  IEEE.

\bibitem[{Uesato et~al.(2019)Uesato, Alayrac, Huang, Stanforth, Fawzi, and
  Kohli}]{uesato2019labels}
Uesato, J.; Alayrac, J.-B.; Huang, P.-S.; Stanforth, R.; Fawzi, A.; and Kohli,
  P. 2019.
\newblock Are labels required for improving adversarial robustness?
\newblock \emph{arXiv preprint arXiv:1905.13725} .

\bibitem[{Wang, Shi, and Osher(2019)}]{wang2019resnets}
Wang, B.; Shi, Z.; and Osher, S. 2019.
\newblock Resnets ensemble via the feynman-kac formalism to improve natural and
  robust accuracies.
\newblock In \emph{Advances in Neural Information Processing Systems},
  1657--1667.

\bibitem[{Wong, Rice, and Kolter(2019)}]{wong2019fast}
Wong, E.; Rice, L.; and Kolter, J.~Z. 2019.
\newblock Fast is better than free: Revisiting adversarial training.
\newblock In \emph{International Conference on Learning Representations}.

\bibitem[{Xie et~al.(2018)Xie, Wang, Zhang, Ren, and
  Yuille}]{xie2018mitigating}
Xie, C.; Wang, J.; Zhang, Z.; Ren, Z.; and Yuille, A. 2018.
\newblock Mitigating Adversarial Effects Through Randomization.
\newblock In \emph{International Conference on Learning Representations}.

\bibitem[{Yun et~al.(2019)Yun, Han, Oh, Chun, Choe, and Yoo}]{yun2019cutmix}
Yun, S.; Han, D.; Oh, S.~J.; Chun, S.; Choe, J.; and Yoo, Y. 2019.
\newblock Cutmix: Regularization strategy to train strong classifiers with
  localizable features.
\newblock In \emph{Proceedings of the IEEE International Conference on Computer
  Vision}, 6023--6032.

\bibitem[{Zagoruyko and Komodakis(2016)}]{Zagoruyko2016WRN}
Zagoruyko, S.; and Komodakis, N. 2016.
\newblock Wide Residual Networks.
\newblock In \emph{BMVC}.

\bibitem[{Zhai et~al.(2019)Zhai, Cai, He, Dan, He, Hopcroft, and
  Wang}]{zhai2019adversarially}
Zhai, R.; Cai, T.; He, D.; Dan, C.; He, K.; Hopcroft, J.; and Wang, L. 2019.
\newblock Adversarially robust generalization just requires more unlabeled
  data.
\newblock \emph{arXiv preprint arXiv:1906.00555} .

\bibitem[{Zhang et~al.(2016)Zhang, Bengio, Hardt, Recht, and
  Vinyals}]{zhang2016understanding}
Zhang, C.; Bengio, S.; Hardt, M.; Recht, B.; and Vinyals, O. 2016.
\newblock Understanding deep learning requires rethinking generalization.
\newblock \emph{arXiv preprint arXiv:1611.03530} .

\bibitem[{Zhang et~al.(2018)Zhang, Cisse, Dauphin, and
  Lopez-Paz}]{zhang2018mixup}
Zhang, H.; Cisse, M.; Dauphin, Y.~N.; and Lopez-Paz, D. 2018.
\newblock mixup: Beyond Empirical Risk Minimization.
\newblock In \emph{International Conference on Learning Representations}.

\bibitem[{Zhang et~al.(2019)Zhang, Yu, Jiao, Xing, Ghaoui, and
  Jordan}]{zhang2019theoretically}
Zhang, H.; Yu, Y.; Jiao, J.; Xing, E.~P.; Ghaoui, L.~E.; and Jordan, M.~I.
  2019.
\newblock Theoretically principled trade-off between robustness and accuracy.
\newblock \emph{arXiv preprint arXiv:1901.08573} .

\end{thebibliography}

\end{document}